\documentclass{bmvc2k}

\title{RS\textsuperscript{3}-Prune: Read-Sparse, Store-Sparse Token Pruning for Video Object Segmentation}

\addauthor{Avilasha Mandal}{avilasha1112@gmail.com}{1,2}
\addauthor{Sarvesh Shashikumar}
{sarveshshashikumar0908@gmail.com}{2}

\addinstitution{
 Attentive AI
}
\addinstitution{
 Indian Institute\\
of Technology\\
 Delhi
}

\runninghead{Mandal, Shashikumar}{RS\textsuperscript{3}-Prune}

\usepackage[ruled,vlined]{algorithm2e}

\usepackage{graphicx}
\usepackage{amssymb}
\usepackage{amsmath}
\usepackage{booktabs}
\usepackage{multirow}
\usepackage{array}
\usepackage{float}
\usepackage{url}
\usepackage{placeins}

\usepackage{enumitem}
\usepackage{microtype}
\usepackage[normalem]{ulem}
\usepackage{hyperref}

\usepackage{colortbl}
\usepackage[table]{xcolor}

\newcolumntype{C}[1]{>{\centering\arraybackslash}p{#1}}
\newcolumntype{R}{>{\raggedleft\arraybackslash}X}

\newcommand{\method}{RS\textsuperscript{3}-Prune\xspace}
\newcommand{\methodLong}{Read-Sparse, Store-Sparse Token Pruning\xspace}
\providecommand{\xspace}{}
\newcommand{\fps}{\,\textsc{fps}}
\newcommand{\vram}{\,\textsc{vram}}

\newcommand{\Jf}{$\mathcal{J}\&\mathcal{F}$ }
\newcommand{\Jj}{$\mathcal{J}$}
\newcommand{\Ff}{$\mathcal{F}$}

\newcommand{\better}[1]{\textcolor{green!50!black}{#1}}
\newcommand{\tbd}{\textcolor{gray}{\,---\,}}
\newcommand{\imgenc}{$\Phi_\mathrm{enc}$ }
\newcommand{\davis}{DAVIS-17 \cite{davis2017}}
\newcommand{\mose}{MOSE \cite{ding2023mose}}
\newcommand{\sav}{SA-V \cite{sam2}}
\newcommand{\ytvos}{YouTube-VOS \cite{ytvos19}}

\definecolor{best}{RGB}{198,239,206}   
\definecolor{second}{RGB}{226,239,218} 
\definecolor{third}{RGB}{255,242,204}  
\begin{document}

\maketitle

\begin{abstract}
 We introduce \method, a training-free token-pruning
recipe that instantiates as a small set of inference time hooks atop existing video object segmentation (VOS) networks. Modern VOS models have converged on a common, expensive design: an image encoder produces a dense token grid for every frame, and a memory bank accumulates these tokens across all previously processed frames to condition future predictions. As a video grows longer, the resulting token budget governs both per-frame latency and peak GPU memory. Hence these models break on use cases such as --- long-form video or real-time deployment on memory-bounded accelerators. In this work we argue that the right axis along which to compress memory-bank VOS is the token budget itself. \method operates in two precise locations within an arbitrary memory-bank VOS pipeline: at the boundary between the image encoder and the memory-attention readout, where we restrict the queries that participate in the cross-frame attention to only a small, geometrically informed subset; and at the boundary between the memory encoder and the memory bank, where we restrict which tokens are ever permitted to enter the bank to those that lie within the object's spatial extent. Over various established benchmarks,  \method delivers  a mean of $38.8\%$ FPS speedup and $13.1\%$ peak memory reduction, while preserving a competitive $\mathcal{J}\&\mathcal{F}$ compared to the unmodified VOS networks.


\end{abstract}

\section{Introduction}
\label{sec:intro}

Video object segmentation (VOS) is the task of tracking and delineating one or more objects across the frames of a video, producing a per-pixel binary mask for each object at every time step. But VOS now lives in a saturated regime. On datasets like \davis, \sav, \mose, \ytvos and LVOSv2 \cite{hong2025lvos}, the SOTA VOS networks
\cite{xmem,cutie,sam2,stm,carion2025sam3segmentconcepts} hover within a few points of the
$\mathcal{J}\&\mathcal{F}$ ceiling. The interesting question is no longer whether these
networks can segment well on a benchmark video; it is whether they
can be \emph{run} at scale, on long videos, on many simultaneous
streams, and on memory-bounded GPUs without exhausting accelerator
memory or stalling under per-frame \cite{Zhu_2026_WACV}. In those regimes the
bottleneck is not the network's algorithmic capability; it is the
number of tokens that flow through its memory pipeline.

The cause is structural, and it is the same across architectures.
Every modern memory-bank network comprises two heavy stages stacked back to
back \cite{sam2}. An image encoder produces a dense token grid for each frame, as the input resolution used by
prompt-driven VOS networks. A memory encoder then converts each frame's
predicted mask, together with the encoder features, into a
key/value pair, and appends this pair to a memory bank that conditions every subsequent
prediction. The cross-frame attention that fuses the memory bank with the current frame is inherently dependent on interactions between both components: the spatial resolution of the frame, the accumulation of more memory frames over the course of a video. These two factors increase both the computational cost and peak memory consumption. The peak memory
grows with video length, throughput falls with
it, and parallel inference workloads run out of memory long before
they run out of capability.

We take the view that the token budget is the quantity that must be reduced,
and that it must be reduced at two distinct locations within the
pipeline. The first location lies between the image encoder and the
memory-attention readout: it is the gate at which the current
frame's tokens are presented as queries to the memory bank, and it is
where we apply our \emph{read-side} prune. The second location lies
between the memory encoder and the memory bank: it is the gate at
which a frame's encoded tokens are committed to long-term storage,
and it is where we apply our \emph{write-side} prune. The first
asks ``which of the current frame's tokens deserve to read''; the
second asks ``which of the current frame's tokens deserve to be
remembered''. We argue that both are legitimate places to compress,
and that compressing both is what unlocks the deployment regimes
above. We package the combined recipe under the name \method ---
\methodLong --- and instantiate it as a training-free patch on five
published VOS networks.

The contribution of this paper is therefore as follows. We argue
that the token budget of memory-bank VOS, and not the algorithmic
gap between published networks, is the real deployment ceiling for
the current generation of systems. We present \method as a single,
symmetric recipe that compresses the token budget at the two
locations where it actually accumulates, and we mathematically
justify the geometric prior that drives it from first principles
about the L2 energy of encoder tokens. We then measure \method on
five published architectures across five benchmarks of deliberately
different character --- \davis, \mose, \sav, \ytvos, LVOSv2 \cite{hong2025lvos} validation sets ---
and report both its strengths and its limits openly.
\section{Related Work}
\label{sec:related}

\paragraph{Image and video object segmentation.}
The Segment Anything Model (SAM)~\cite{kirillov2023segment} established a new paradigm for image segmentation, training a large vision transformer on a curated dataset of over one billion masks to produce a general-purpose segmentation engine operable via sparse prompts such as points, boxes, or masks. SAM2~\cite{sam2} extended this foundation to video by introducing a streaming memory mechanism: a lightweight memory encoder folds each frame's prediction into a compact representation that is stored in a fixed-size rolling buffer, which a memory attention module then reads to condition segmentation of the next frame. This design preserves SAM's interface while handling temporal coherence, achieving state-of-the-art results across several datasets at the cost of a substantially larger token budget than purpose-built VOS architectures.
\paragraph{Memory bank video object segmentation.}
Early approaches relied on online fine-tuning~\cite{caelles2017one, khoreva2016learning}, adapting a segmentation network to the reference frame at test time — achieving strong accuracy at the cost of prohibitive per-video computation. Propagation-based methods~\cite{oh2018fast, Perazzi_2016_CVPR} sidestepped this by warping masks forward using optical flow or learned correspondences, trading some robustness for dramatically faster inference. The field shifted decisively with Space-Time Memory Networks~\cite{stm}, which replaced explicit propagation with a soft nearest-neighbour readout over a growing memory bank of past frame embeddings, elegantly handling re-identification and long-term appearance change without any test-time optimisation. This blueprint — an image encoder producing a dense per-frame token grid, a memory attention module conditioning the current frame on a bank of past key-value pairs written by a memory encoder, and a mask decoder producing the final segmentation — has since become the dominant structural commitment in the field. Subsequent work refined every component: XMem~\cite{xmem} introduced a multi-granularity memory hierarchy to manage bank growth over long videos; Cutie~\cite{cutie} replaced per-pixel readout with object-level tokens to suppress background noise; and SAM2~\cite{sam2} scaled the paradigm to a large vision foundation model with streaming memory. Despite their differences in architecture and scale, all of these methods share the same growing-bank cross-attention bottleneck, and it is precisely this shared bottleneck — the dense token grid attending into an ever-growing memory — that \method targets, without committing to any one design.
\paragraph{Token pruning.}
Token pruning has a long history in dense vision. DynamicViT
\cite{tokenprune-imagenet} and SparseViT \cite{sparsevit}
selectively drop tokens from a vision transformer's forward pass on
the basis of attention scores or token energies; EViT
\cite{liang2022patchesneedexpeditingvision} and Evo-ViT
\cite{xu2021evovitslowfasttokenevolution} similarly reorganise or
evolve tokens using class-token attention, while SPViT
\cite{kong2022spvitenablingfastervision} learns a soft,
latency-aware token selector. A parallel line merges rather than
discards redundant tokens: ToMe \cite{bolya2023tokenmergingvitfaster}
performs bipartite soft matching between similar tokens, and PPT
\cite{wu2024ppttokenpruningpooling} and M2M-TAG \cite{m2mtag}
generalise pruning and merging into a single pooling-based or
many-to-many transformation, an idea further unified by Token
Transforming \cite{zeng2025tokentransformingunifiedtrainingfree}.
VLTP \cite{vltp}, \cite{mandal2025fastsam2textdriventoken} recently brought token reduction into prompted
segmentation by ranking tokens against a text-derived prior. To our
knowledge all of this work --- whether it prunes, merges, or
transforms tokens --- operates exclusively at the attention-time
query boundary. The contribution of \method is to identify a
second, structurally distinct location at which token reduction is
both possible and useful for video object segmentation --- the
memory-write boundary --- and to show that acting at both locations
is what unlocks the deployment regimes we care about.
\section{Method}
\label{sec:method}


\begin{figure*}[!htb] 
	\centering
	\includegraphics[width=\linewidth] {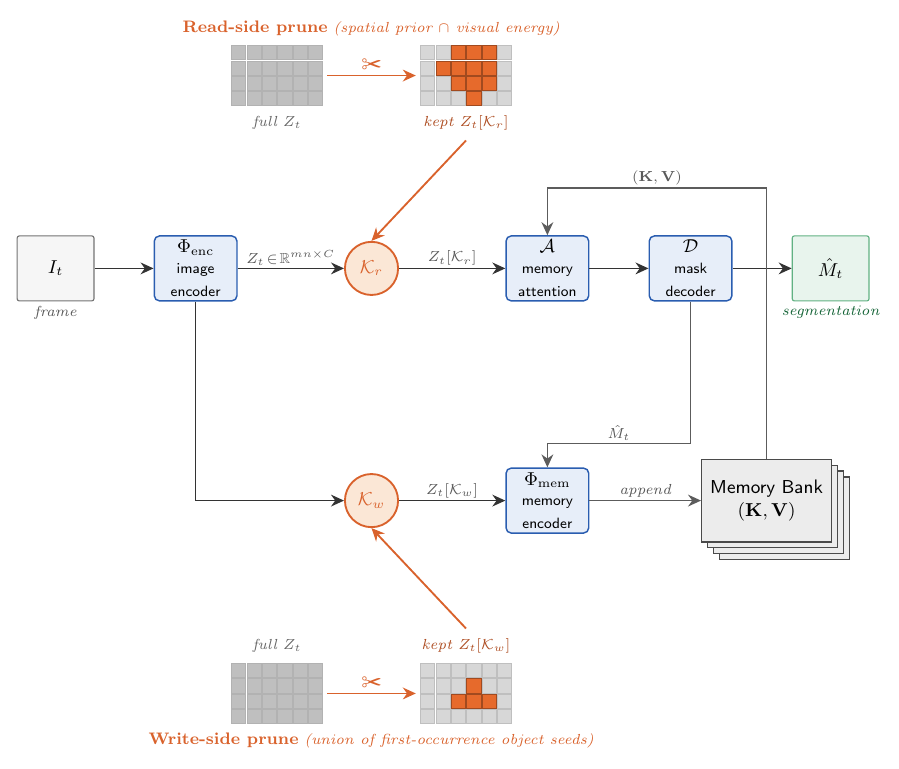} 
	\caption{\textbf{Overview of \method.} The standard memory-        bank VOS
        pipeline (blue) passes all $mn$ encoder tokens through memory
        attention $\mathcal{A}$ and into the memory encoder $\Phi_{\mathrm{mem}}$.
        \method inserts two training-free pruning operators (orange) at the
        two token boundaries. The \emph{read-side} prune $\mathcal{K}_r$
        retains only tokens that lie within the spatial prior and carry
        high visual energy, reducing the query load on $\mathcal{A}$.
        The \emph{write-side} prune $\mathcal{K}_w$ retains only tokens
        anchored to first-occurrence object seeds before they are written
        to the memory bank $(\mathbf{K}, \mathbf{V})$, controlling bank
        growth. Both pruning steps are applied to the frozen pretrained
        checkpoint with no retraining.}
	\label{fig:architecture}
\end{figure*}

\begin{figure*}[!htb] 
	\centering
	\includegraphics[width=\linewidth] {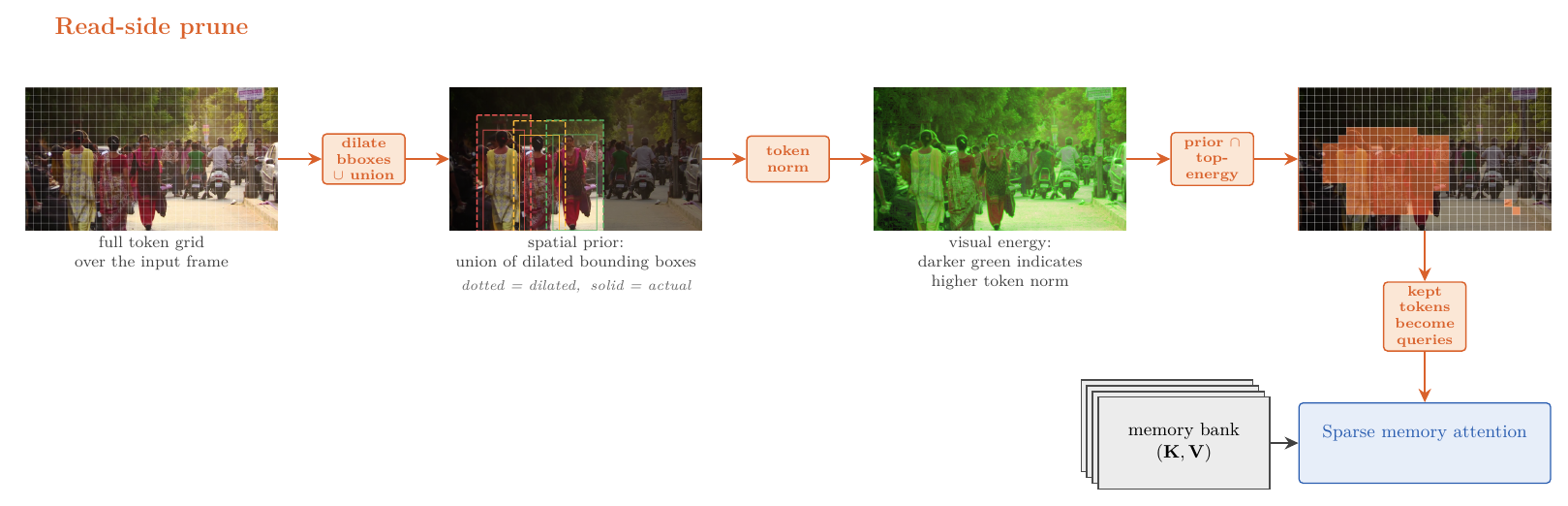} 
	\caption{\textbf{Read-side prune.} Starting from the full $mn$ token grid, bounding boxes from the first-frame annotation are dilated and unioned to form a spatial prior. Token norms are computed as a proxy for visual energy, with higher-norm tokens (darker green) indicating visually salient regions. The final kept set $\mathcal{K}_r$ is the intersection of the spatial prior with the top-energy tokens; only these tokens participate as queries in the subsequent sparse memory attention over $(\mathbf{K}, \mathbf{V})$.}
	\label{fig:readside_prune}
\end{figure*}

We treat a memory-bank video object segmentation pipeline as a chain of four well-defined stages. An image encoder \imgenc maps the current frame $I_t$ (the RGB image at time step $t$) into a dense token grid $Z_t \in \mathbb{R}^{mn \times C}$, where $m \times n$ is the spatial grid resolution and $C$ is the feature channel dimension. A memory attention module $\mathcal{A}$ conditions $Z_t$ on a memory bank $(\mathbf{K}, \mathbf{V})$ accumulated from previous frames, where $\mathbf{K} \in \mathbb{R}^{N_\mathrm{mem} \times C}$ and $\mathbf{V} \in \mathbb{R}^{N_\mathrm{mem} \times C}$ are the stored key and value matrices, $N_\mathrm{mem}$ is the total number of memory tokens. The memory encoder $\Phi_\mathrm{mem}$, run every $T$ frames, folds the current frame's prediction and tokens into a fresh key/value pair and appends it to the bank. A mask decoder $\mathcal{D}$ finally turns the conditioned representation into the per-frame segmentation.


As shown in figure \ref{fig:architecture}, \method makes two interventions in this chain, at the two places where the token budget enters or leaves the long-lived memory bank. The first intervention sits at the boundary \emph{between the image encoder and the memory attention}, and decides which of the current frame's $mn$ tokens are allowed to participate as queries in the cross-frame readout. We call this the \emph{read-side} prune (Sec.~\ref{sec:read-side}). The second intervention sits at the boundary \emph{between the memory encoder and the memory bank}, and decides which of the current frame's tokens are permitted to enter the bank in the first place. We call this the \emph{write-side} prune (Sec.~\ref{sec:write-side}). Both interventions keep the backbone, the memory module, and the mask decoder unchanged, and they run on the original pretrained checkpoint: there is no retraining, no fine-tuning, and no new parameter.

\subsection{Read-side pruning: a per-frame, per-object spatial gate}
\label{sec:read-side}

We install a spatial gate around \imgenc of the VOS network so that, for
each frame, only the tokens \emph{around the tracked object(s)} pay
the full memory-attention cost; everywhere else either contributes
zero at read time, or is never written into the memory bank at write time
(Sec.~\ref{sec:write-side}). The gate is driven by a per-frame,
per-object bounding box derived from ground-truth first-frame mask under the semi-supervised protocol. From the second frame onward the
gate is propagated temporally using model's own previous-frame
predictions, with no additional supervision. This is shown in figure \ref{fig:readside_prune}.
\paragraph{Seeding the spatial prior into frame 0.}
For each object identifier $o \in \mathcal{O}$ we obtain an initial
binary mask $M_o^{(0)} \in \{0,1\}^{m \times n}$. In the
semi-supervised protocol $M_o^{(0)}$ is read directly from the first-frame palette mask. From each $M_o^{(0)}$ we compute its tight bounding
box $\mathcal{B}_o^{(0)} = \mathrm{bbox}(M_o^{(0)})$. In VOS architectures like SAM~2 \cite{sam2} which
expects a seed prior, re-uses this same bounding box as seed prompt for it's prompt encoder.

\paragraph{From a pixel-space box to a token-grid prior.}
\imgenc of the VOS network emits a feature map of shape $(1, C, m, n)$
with $mn$ tokens and $C$ feature vector dimension at stride
$s$. We project each seed box $\mathcal{B}_o^{(0)}$ onto this
$m \times n$ grid by nearest-neighbour downsampling (pixel
coordinates divided by $s$), and dilate it by $r_s$ grid cells
with an $(r_s-1) \times (r_s-1)$ structuring element to absorb the quantisation
error introduced by the down-projection. The seed spatial prior is
the union of the per-object dilated grids:
\begin{equation}
\mathcal{P}^{(0)} \;=\;
  \bigcup_{o \in \mathcal{O}}
  \mathrm{dilate}_{r_s}\!\bigl(
    \mathrm{down}_{s}(\mathcal{B}_o^{(0)})
  \bigr)
  \;\subset\; \{1,\dots,m\}\times\{1,\dots,n\}.
\label{eq:seed-prior}
\end{equation}

\paragraph{Temporal propagation of the prior.}
For frames $t \ge 1$ we already have a multi-object prediction
$\hat{M}_{t-1}$ from the previous step. We rebuild the prior at $t$
by applying the same project-and-dilate operation to the
\emph{predicted} per-object masks of the previous frame:
\begin{equation}
\mathcal{P}_o^{(t)} \;=\;
  \mathrm{dilate}_{r_s}\!\bigl(
    \mathrm{down}_{s}\bigl(\hat{M}_{t-1}|_o\bigr)
  \bigr),
\qquad
\mathcal{P}^{(t)} \;=\; \bigcup_{o \in \mathcal{O}} \mathcal{P}_o^{(t)}.
\label{eq:dyn-prior}
\end{equation}
The seed mask is therefore only consulted at $t = 0$; from $t = 1$
onward the gate follows model's own predictions. There is no second
supervisory signal: the prior is the geometric trace the tracker
has produced for itself.

\paragraph{We retain a fraction of the prior.}
After dilation, $\mathcal{P}^{(t)}$ typically covers only $30$--$50\%$ of
the total token count, which is a useful reduction but still loose:
not every cell inside an object's dilated box is equally
informative. We further restrict the keep-set to the cells whose
encoder tokens carry the most visual energy. Let $z_i \in
\mathbb{R}^{C}$ be the encoder token at grid cell
$i \in \mathcal{P}^{(t)}$, and let $\alpha_i \propto
\exp(\mathbf{q}^\top W_K z_i / \sqrt{C})$ be its attention weight
under a query $\mathbf{q}$. The Cauchy--Schwarz inequality bounds
token $i$'s contribution to the attention readout:
\begin{equation}
\| \alpha_i \, W_V z_i \|_2 \;\le\;
  \alpha_i \cdot \| W_V \|_2 \cdot \| z_i \|_2.
\label{eq:cs-bound}
\end{equation}
So under a fixed query a token can influence the readout only in
proportion to its norm. The norm $\|z_i\|_2$ in turn tracks the
encoder's response to local image content: textured object regions
(edges, contrast, high-frequency structure) drive $\|z_i\|_2$
\emph{up}, while smooth uniform regions (sky, walls, out-of-focus
background) drive it \emph{down}. This is the ``visual-energy''
empirical regularity that token pruning already exploits in
classification~\cite{tokenprune-imagenet,sparsevit}, and it is the
reason the geometric prior and the energy ranking agree: object
pixels generate, on average, larger token energies than background
pixels. We accordingly rank the tokens in the prior by $\|z_i\|_2^2$
and keep only the top $\mathcal{K}_r^{(t)}$ tokens ($\le\ \rho \cdot mn$):
\begin{equation}
\mathcal{K}_r^{(t)} \;=\;
  \underset{i \in \mathcal{P}^{(t)}}{\mathrm{arg\text{-}top}_{\rho mn}}
  \, \|z_i\|_2^2
\label{eq:read-keepset}
\end{equation}
The geometric prior supplies the \emph{spatial} support (object vs.\
background); the energy ranking inside it supplies the
\emph{semantic} support (textured vs.\ smooth). Tokens that pass
both filters are the ones a memory readout can actually use; the
rest are bounded by Eq.~\ref{eq:cs-bound} and can be dropped at
near-zero cost in $\mathcal{J}\&\mathcal{F}$.

\paragraph{Disappearance recovery via a streak-counted bounding box.}
Real videos contain objects that occlude, leave the frame, and
re-enter. Whenever the previous prediction $\hat{M}_{t-1}|_o$ is
empty, Eq.~\ref{eq:dyn-prior} would produce an empty prior for $o$,
and the read keep-set could then never include the cells the object
has to re-emerge into. We mitigate this with a
\emph{streak-counted bounding-box fallback}. For each object we
maintain two pieces of state across the video: the last non-empty
bounding box $\hat{\mathcal{B}}_o = \mathrm{bbox}\bigl(
\hat{M}_\tau|_o\bigr)$ observed at its most recent successful frame
$\tau < t$, and a streak counter $\Delta_o$ that increments by one
for every consecutive frame on which $\hat{M}|_o$ has been empty.
When $\hat{M}_{t-1}|_o$ is empty, the fallback prior is
\begin{equation}
\mathcal{P}_o^{(t)} \;=\;
  \mathrm{dilate}_{\,r_s \,+\, \min(\Delta_o,\, c)}\!\bigl(
    \mathrm{down}_s(\hat{\mathcal{B}}_o)
  \bigr),
\qquad
\label{eq:recovery}
\end{equation}
i.e.\ the cached box is dilated more aggressively the longer the
object has been missing, up to a cap of $r_s + c$ grid cells. This
cap $c$ exists because an indefinitely growing prior would defeat the
purpose of pruning. As soon as a frame produces a non-empty mask for
$o$, the streak resets to $\Delta_o = 0$ and Eq.~\ref{eq:dyn-prior}
resumes. Stably absent objects are therefore searched across up to
$c$ frames of growing radius before the recovery saturates; we
never declare an object permanently absent, so $\hat{\mathcal{B}}_o$
persists for the whole video.

\paragraph{Sparse memory attention.}
The VOS model's memory attention is a stack of $L$ self-/cross-attention
layers that take, at each call, the full $mm$ encoder tokens
as queries and the running memory bank $(\mathbf{K}, \mathbf{V})$ as
keys and values. We patch this stack at every layer's entry: the
queries are index-gathered onto the $|\mathcal{K}_r^{(t)}|$ kept
positions, $\tilde{\mathbf{q}}_t = \mathbf{q}_t[\mathcal{K}_r^{(t)}]$,
the attention runs on these queries only, and the result
is scattered back into a length-$mn$ tensor with zeros at the
dropped positions:
\begin{equation}
\mathcal{A}_{\mathrm{sparse}}(\mathbf{q}_t \mid \mathbf{K}, \mathbf{V})
  \;=\;
  \mathcal{S}_{\mathcal{K}_r^{(t)}}\!\Bigl(
    \mathrm{softmax}\!\bigl(
      \tilde{\mathbf{q}}_t \mathbf{K}^\top / \sqrt{C_k}
    \bigr) \, \mathbf{V}
  \Bigr).
\label{eq:read-attn}
\end{equation}
After scatter, the attention output is still a $m \times n$
token grid --- the model-native shape the mask decoder expects ---
with $|\mathcal{K}_r^{(t)}|$ non-zero entries placed at their
original $(x,y)$ coordinates and zeros elsewhere. The mask decoder
and every downstream tensor shape are therefore identical to the
dense pipeline; only the work performed by memory attention shrinks
heavily.

\paragraph{RoPE on the sparse subset.}
The model's memory attention uses rotary positional
embeddings~\cite{rope} with a precomputed frequency table $f \in
\mathbb{C}^{mn \times C/2}$, indexed by the position of each query
in the full $m \times n$ grid. If we left the standard
implementation untouched, the gathered queries
$\tilde{\mathbf{q}}_t$ would silently be paired with the
\emph{prefix} $f[\,:\,|\mathcal{K}_r^{(t)}|]$ of the table, which
is incorrect: the kept indices are $not$ contiguous, and the prefix
slice encodes the wrong $(x,y)$ positions. We replace the slice by
an index-gather that mirrors the query gather of
Eq.~\ref{eq:read-attn}:
\begin{equation}
\tilde{f} \;=\; f[\mathcal{K}_r^{(t)}],
\label{eq:rope-gather}
\end{equation}
so that every gathered query is rotated by the frequency that
corresponds to its \emph{original} grid cell. This single
correction is what makes sparse memory attention numerically
equivalent to dense memory attention in the limit
$\rho \rightarrow 1$: at $\rho = 1$ the gather and scatter become
identity maps and Eq.~\ref{eq:read-attn} reduces exactly to the
unmodified memory readout.

\subsection{Write-side pruning: which tokens enter the bank}
\label{sec:write-side}

\begin{figure*}[!htb] 
	\centering
	\includegraphics[width=\linewidth] {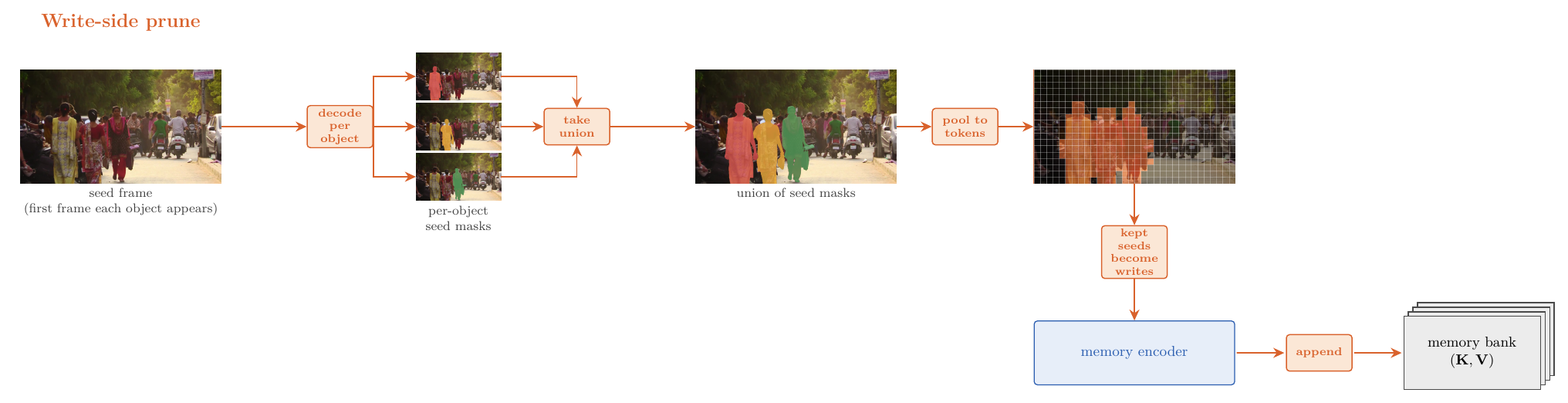} 
	\caption{\textbf{Write-side prune.} For each seed frame — the first frame in which an object appears — per-object masks are decoded and unioned into a single seed mask. This union is pooled to the token grid to yield the kept index set $\mathcal{K}_w$, which selects the tokens passed to the memory encoder $\Phi_{\mathrm{mem}}$. Only these seed-anchored tokens are appended to the memory bank $(\mathbf{K}, \mathbf{V})$, preventing background clutter from accumulating in the bank across frames.}
	\label{fig:writeside_prune}
\end{figure*}

\paragraph{Why a write-side prune is needed.}
Read-side pruning compresses the query axis of cross-frame
attention but leaves $N_\mathrm{mem}$ untouched: every memory
commit still deposits a full $mn$-token slab into the memory bank, and on
a long video this is what eventually exhausts GPU memory. Write-side
pruning attacks the same problem at its source by deciding which
tokens are admitted into the bank in the first place. Our write-side pruning is displayed in figure \ref{fig:writeside_prune}.

\paragraph{Object-boundary keep-set on the encoder grid.}
The object boundary that defines the write keep-set is read
directly from the seed mask: for each object $o \in \mathcal{O}$ we
take $M_o^{(\tau_o)}$, the binary mask of $o$ at its
first-appearance frame $\tau_o$, downsample it by nearest-neighbour
to the encoder stride $s$, and dilate the result by $D_w$ grid
cells. The write keep-set is the union over objects of these
per-object dilated regions:
\begin{equation}
\mathcal{K}_w \;=\;
  \bigcup_{o \in \mathcal{O}}
  \mathrm{dilate}_{D_w}\!\bigl(
    \mathrm{down}_s\bigl(M_o^{(\tau_o)}\bigr)
  \bigr),
\qquad |\mathcal{K}_w| \;\ll\; mn.
\label{eq:write-keepset}
\end{equation}
Geometrically, $\mathcal{K}_w$ is a strict superset of the spatial
cells the seed object occupies, plus a margin proportional to $D_w$
that anticipates the object's motion envelope across the rest of
the video. Letting $\tilde{Z}_t = Z_t[\mathcal{K}_w]$ denote the
encoder-output tokens gathered at the write keep-set, only
$\tilde{Z}_t$ is passed to the memory encoder $\Phi_\mathrm{mem}$,
and only its output is appended to the bank $(\mathbf{K},
\mathbf{V})$. Every stored frame is therefore $|\mathcal{K}_w|$
tokens wide rather than $mn$, and the bank grows at a strictly
smaller per-frame stride.

\paragraph{Why this is safe.}
Two properties of $\mathcal{K}_w$ make the write-side prune safe.
First, $\mathcal{K}_w$ is computed once at $\tau_o$ and held fixed
for every subsequent write of the same sequence: this uniform
per-frame stride is what lets the memory bank's long-term
consolidation, sieving, and prototype-potentiation operations continue working without modification, since
they slice the bank by a constant per-frame length and we provide
one. Second, $\mathcal{K}_w$ is geometrically aligned with the
read-side keep-set $\mathcal{K}_r^{(t)}$ of
Eq.~\ref{eq:read-keepset}: the read side has no reason to query
positions whose bank entries are empty, and the relation
$\mathcal{P}^{(t)} \subseteq \mathcal{K}_w$ (which holds whenever
$D_w \ge r_s + c$) guarantees that every active query position has
a stored counterpart. On the few sequences in which the seed
footprint covers $> 95\%$ or $< 5\%$ of $mn$, write-side pruning
falls through to the original write path so that
almost-full-frame or almost-empty objects are not lost.

\subsection{Per-frame morphological closure}
\label{sec:morph}

The scattered sparse readout (Eq.~\ref{eq:read-attn}) places zeros
at non-kept cells, which the mask decoder occasionally reads as
salt-and-pepper holes inside otherwise contiguous object interiors.
We close these holes with a single per-object morphological closure
applied to each predicted binary mask $\hat{M}_t|_o$ using a $9
\times 9$ square structuring element. Closure is a dilation followed
by an erosion: it fills holes smaller than the structuring element
and re-knits thin interior gaps, but does not grow the outer
boundary of the object. The closure runs at the decoder's output
resolution, and only after
the closure do we take the per-pixel argmax across objects to
produce the final multi-object palette mask $\hat{M}_t$. The cost
of one $9 \times 9$ closure per object per frame is negligible
compared to memory attention.

\subsection{Why the geometric prior works: token energy on the encoder grid}
\label{sec:why}

The read-side and write-side keep-sets are derived from the object
mask, not from the encoder features. One could plausibly object
that a feature-driven prior --- for instance, retaining tokens by
$\ell_2$ norm or by their contribution to the attention readout
--- would carry the same information without the need for an
explicit mask. The Cauchy--Schwarz argument in
Eq.~\ref{eq:cs-bound} is the reason a geometric prior is, in fact,
a strict refinement of these feature-driven priors: in the limit of
a perfect mask, the spatial prior selects exactly the cells whose
tokens carry the high $\ell_2$ energy, without having to threshold
a noisy scalar. In our setting we do not have a perfect mask --- we
have the previous frame's prediction, which is a reasonable but
noisy proxy --- and the dilation $r_s$ absorbs the residual
uncertainty. The result is that retaining $\mathcal{K}_r^{(t)} \cup
\mathcal{K}_w$ keeps the tokens whose energy is, in expectation,
the largest, and discards the ones whose energy is, in expectation,
near zero.

\subsection{Cross-Architecture Portability}
\label{sec:portability}

\method applies cleanly to any memory-bank video object segmentation
network whose pipeline exposes the two boundaries identified in
Sec.~\ref{sec:method}: a memory-attention readout into which the
current frame presents queries, and a memory-bank insertion path at
which the current frame's tokens are committed to long-term storage.

\section{Experiments}
\label{sec:experiments}

\subsection{Datasets and Implementation Details}
We evaluate \method on five standard video-object-segmentation benchmarks: \davis (evaluated using davis 2017 per object evaluation protocol),
\sav (evaluated using Meta's official evaluation script),
\ytvos (evaluated using davis 2017 per object evaluation protocol),
\mose (evaluated using Codabench evaluation script) and LVOS-v2~\cite{hong2025lvos},
whose 140 validation clips average $\sim$1.14 minutes, roughly $5\times$
longer than DAVIS-17 and specifically stress-test the long-video
deployment regime motivating this work, rather than serving as a fifth
short-clip peer to the other four. All scores are measured on a single NVIDIA RTX-4090-class GPU using a batch size of 1.

\subsection{Main result across five benchmarks}
Table~\ref{tab:main} reports $\mathcal{J}$, $\mathcal{F}$, $\mathcal{J}\&\mathcal{F}$, \fps{} and \vram{} for the five memory-bank architectures (STM \cite{stm}, XMem \cite{xmem}, Cutie \cite{cutie}, SAM2 \cite{sam2}, SAM3.1 \cite{carion2025sam3segmentconcepts}) under both the Vanilla and the full \method configuration on all five datasets. The \textbf{Vanilla} column corresponds to each network's published checkpoint with no modification, meanwhile the \textbf{\method} column uses the same checkpoint with the read-side keep-set at $\rho{=}0.3$ and the write-side keep-set at the network-specific dilation. No retraining is performed.

We evaluate each model along three axes. Region similarity $\mathcal{J}$ measures the intersection-over-union between the predicted and ground-truth masks, contour accuracy $\mathcal{F}$ measures the alignment of their boundaries, and we report their mean \Jf as the primary accuracy metric. Inference speed is reported in frames per second (FPS), measured on a single GPU. Peak GPU memory consumption is reported in megabytes (MB) and captures the maximum VRAM allocated during inference.

\begin{table*}[t]
\setlength{\abovecaptionskip}{3pt}
\centering \footnotesize
\setlength{\tabcolsep}{3pt}
\setlength{\heavyrulewidth}{0.25em}
\renewcommand{\arraystretch}{1.0}
\begin{tabular}{ll|ccccc|ccccc}
\toprule
& & \multicolumn{5}{c|}{Vanilla} & \multicolumn{5}{c}{\method} \\
Dataset & Model
  & \Jj{} & \Ff{} & \Jf{} & \fps & \vram{}
  & \Jj{} & \Ff{} & \Jf{} & \fps & \vram{} \\
\midrule
\multirow{5}{*}{\davis}
  & STM~\cite{stm}                              & 0.775 & 0.828 & 0.802 & 37.6 & 1043 & 0.757 & 0.782 & 0.770 & \textbf{49.4} & \textbf{728}  \\
  & XMem~\cite{xmem}                             & 0.829 & 0.896 & 0.862 & 37.2 & 845  & 0.821 & 0.876 & 0.849 & \textbf{69.7} & \textbf{655}  \\
  & Cutie~\cite{cutie}                           & 0.847 & 0.911 & 0.879 & 23.8 & 436  & 0.841 & 0.895 & 0.868 & \textbf{75.6} & \textbf{424}  \\
  & SAM2~\cite{sam2}                             & 0.858 & 0.906 & 0.882 & 9.6  & 1809 & 0.774 & 0.819 & 0.797 & \textbf{11.4} & \textbf{1791} \\
  & SAM3.1~\cite{carion2025sam3segmentconcepts}  & 0.529 & 0.590 & 0.560 & 8.70 & 7375 & 0.517 & 0.579 & 0.548 & \textbf{8.70} & \textbf{6154} \\
\midrule
\multirow{5}{*}{\sav}
  & STM~\cite{stm}                              & 0.474 & 0.538 & 0.506 & 32.2 & 1327 & 0.489 & 0.516 & 0.502 & \textbf{50.5} & \textbf{739}  \\
  & XMem~\cite{xmem}                             & 0.575 & 0.651 & 0.613 & 65.8 & 859  & 0.513 & 0.569 & 0.541 & \textbf{71.9} & \textbf{645}  \\
  & Cutie~\cite{cutie}                           & 0.607 & 0.671 & 0.639 & 74.1 & 501  & 0.572 & 0.620 & 0.596 & \textbf{78.0} & \textbf{428}  \\
  & SAM2~\cite{sam2}                             & 0.424 & 0.505 & 0.465 & 9.1  & 2317 & 0.481 & 0.532 & \better{0.506} & \textbf{10.8} & \textbf{2243} \\
  & SAM3.1~\cite{carion2025sam3segmentconcepts}  & 0.457 & 0.479 & 0.468 & 3.7  & 8543 & 0.420 & 0.437 & 0.429 & \textbf{6.3}  & \textbf{8450} \\
\midrule
\multirow{5}{*}{\ytvos}
  & STM~\cite{stm}                              & \tbd  & \tbd  & \tbd  & \tbd  & \tbd  & \tbd  & \tbd  & \tbd  & \tbd  & \tbd  \\
  & XMem~\cite{xmem}                             & 0.819 & 0.863 & 0.841 & 68.8 & 721  & 0.791 & 0.831 & 0.811 & \textbf{69.5} & \textbf{650}  \\
  & Cutie~\cite{cutie}                           & 0.797 & 0.839 & 0.818 & 73.9 & 486  & 0.759 & 0.798 & 0.779 & \textbf{74.8} & \textbf{431}  \\
  & SAM2~\cite{sam2}                             & 0.586 & 0.590 & 0.588 & 10.3 & 1282 & 0.626 & 0.645 & \better{0.636} & \textbf{10.8} & \textbf{1267} \\
  & SAM3.1~\cite{carion2025sam3segmentconcepts}  & 0.497 & 0.515 & 0.506 & 6.9  & 8010 & 0.465 & 0.478 & 0.471 & \textbf{7.2}  & \textbf{7500} \\
\midrule
\multirow{5}{*}{\mose}
  & STM~\cite{stm}                              & 0.416 & 0.490 & 0.454 & 56.2 & 1240 & 0.438 & 0.483 & \better{0.461} & \textbf{60.4} & \textbf{610}  \\
  & XMem~\cite{xmem}                             & 0.527 & 0.609 & 0.569 & 66.2 & 924  & 0.457 & 0.539 & 0.498 & \textbf{69.9} & \textbf{788}  \\
  & Cutie~\cite{cutie}                           & 0.657 & 0.739 & 0.699 & 71.9 & 510  & 0.630  & 0.708  & 0.669 & \textbf{74.4} & \textbf{490}  \\
  & SAM2~\cite{sam2}                             & 0.373 & 0.424 & 0.398 & 8.7  & 1627 & 0.462 & 0.541 & \better{0.502} & \textbf{11.2} & \textbf{1591} \\
  & SAM3.1~\cite{carion2025sam3segmentconcepts}  & 0.354 & 0.386 & 0.370 & 5.8  & 8108 & 0.308 & 0.331 & 0.319 & \textbf{6.5}  & \textbf{7950} \\
\midrule
\multirow{5}{*}{LVOS-v2~\cite{hong2025lvos}}
  & STM~\cite{stm}                              & 0.396 & 0.324 & 0.360 & 6.9  & 17542 & 0.245 & 0.255 & 0.250 & \textbf{16.2} & \textbf{17143} \\
  & XMem~\cite{xmem}                             & 0.514 & 0.534 & 0.524 & 31.9 & 1074  & 0.564 & 0.576 & \better{0.570} & \textbf{40.9} & \textbf{989}   \\
  & Cutie~\cite{cutie}                           & 0.697 & 0.673 & 0.685 & 30.1 & 1323  & 0.633 & 0.619 & 0.626 & \textbf{52.6} & \textbf{1277}  \\
  & SAM2~\cite{sam2}                             & 0.769 & 0.769 & 0.769 & 7.8  & 15197 & 0.650 & 0.670 & 0.660 & \textbf{13.2} & \textbf{12961} \\
  & SAM3.1~\cite{carion2025sam3segmentconcepts}  & 0.317 & 0.497 & 0.407 & 5.2  & 25329 & 0.370 & 0.372 & 0.371 & \textbf{6.3}  & \textbf{24563} \\
\bottomrule
\end{tabular}
\caption{\textbf{Main results of \method across five VOS benchmarks.} Consistent with SOTA VOS evaluation metrics, we report $\mathcal{J}$, $\mathcal{F}$ and the \Jf score, and to display the efficiency and lower resources utilised, we report the FPS and VRAM used. VRAM is measured in \texttt{MB}. \better{Green} marks a gain. \tbd{} indicates that STM cannot handle new object emergence in middle of the video sequence, a characteristic of YT-VOS. \textbf{bold} \fps{}/\vram{} in the \method{} block marks improvement over Vanilla}
\label{tab:main}
\end{table*}


Averaged across the $all$ architectures and datasets with measured
timings in Table~\ref{tab:main}, \method delivers a mean
$\textbf{38.8\% FPS speedup}$ and a mean $\textbf{13.1\% peak VRAM reduction}$
relative to the unmodified networks. The gain is
positive per architecture: averaged over five benchmarks (considering just four datasets for STM as STM has no released semi-supervised YT-VOS runner), \method
gives $+58\%$ FPS,$-32\%$ VRAM on STM,
$+26\%$ FPS,$-16\%$ VRAM on XMem,
$+60\%$ FPS,$-7\%$ VRAM on Cutie,
$+28\%$ FPS,$-4\%$ VRAM on SAM2,
$+22\%$ FPS,$-6\%$ VRAM on SAM3.1.

\Jf{}-remains competitive over datasets and architectures, and on several video sequences it is \emph{better} than vanilla despite attending to only a
fraction of the tokens and smaller context. This is due to the fact of keeping only the geometrically-supported foreground tokens,
the read-side prune discards background context that a full-softmax
readout would otherwise integrate as noise. On long or sparsely-annotated
video (SAM2 on \sav{}, YT-VOS and \mose{}) the prune therefore acts as a
denoiser, lifting \Jf{} by several points while still running faster and
lighter. Concretely, at $\rho{=}0.3$ SAM2's per-frame memory-attention
query budget drops from the full $64{\times}64$ grid ($4096$ tokens) to
$\approx\!1228$ ($3.3\times$ fewer), yet \Jf{} does not fall --- it
improves --- because the retained tokens are precisely the
object-relevant ones.

\paragraph{No human in the loop during the run.}
All numbers above come from \method's token pruning technique
alone: we do \emph{not} issue any human-in-the-loop prompt between frames
of a sequence; the mask is propagated from the seed annotation exactly as
in each backbone's standard VOS protocol. The reported \fps{}/\vram{}
gains and the competitive \Jf{} are thus attributable purely to
the read- and write-side pruning. Neither have we leveraged human in the loop prompting for the unpruned vanilla VOS network for fair and equal comparison. Because the prune is orthogonal to
prompting, this same efficiency profile carries over unchanged when an
interactive prompt is later added to both the pruned and unpruned pipeline.

\subsection{Comparison of \method{} against
prior efficient-VOS methods} We report the comparison of \method{} against
other efficient-VOS methods cited in prior works --- SparseViT~\cite{sparsevit},
DynamicViT~\cite{tokenprune-imagenet}, and VLTP~\cite{vltp} --- at $\rho{=}0.30$, these methods incur huge loss in accuracy when plugged in with vanilla VOS methods. Inherently targeting an efficiency atop image segmentation backbones these methods are not competitive in video settings where the dynamicity of objects is a concern. Without a video/object prior they retain the background and drop the tracked object's (foreground) tokens; \method's geometric prior avoids this underperformance.(Table~\ref{tab:baselines})

\begin{table*}[t]
\setlength{\abovecaptionskip}{3pt}
\centering \footnotesize
\setlength{\tabcolsep}{3pt}
\setlength{\heavyrulewidth}{0.25em}
\renewcommand{\arraystretch}{1.05}
\begin{tabular}{ll|ccccc}
\toprule
Dataset & Method & SAM2 & SAM3.1 & STM & XMem & Cutie\\
\midrule
\multirow{5}{*}{\davis}
 & \raisebox{0.5\height}{Vanilla} & \shortstack{.882\\9.6/1809}  & \shortstack{.560\\8.7/7375}  & \shortstack{.802\\37.6/1043} & \shortstack{.862\\37.2/845}  & \shortstack{.879\\23.8/436}\\
 \cmidrule{2-7}
 & \raisebox{0.5\height}{SparseViT~\cite{sparsevit}}  & \shortstack{.508\\12.4/2324} & \shortstack{.545\\8.5/7175} & \shortstack{.647\\22.5/1730} & \shortstack{.791\\46.0/1207} & \shortstack{.808\\80.2/638}\\
 \cmidrule{2-7}
 & \raisebox{0.5\height}{DynamicViT~\cite{tokenprune-imagenet}} & \shortstack{.041\\12.5/2324} & \shortstack{.392\\8.5/7175} & \shortstack{.612\\44.8/1730} & \shortstack{.554\\68.7/1207} & \shortstack{.798\\33.1/638}\\
 \cmidrule{2-7}
 & \raisebox{0.5\height}{VLTP~\cite{vltp}}       & \shortstack{.133\\12.1/2905} & \shortstack{.012\\8.5/7175} & \shortstack{.275\\45.5/1730} & \shortstack{.291\\74.8/1207} & \shortstack{.553\\32.5/638}\\
 \cmidrule{2-7}
 & \raisebox{0.5\height}{\textbf{\method}} & \shortstack{\textbf{.797}\\11.4/1791} & \shortstack{\textbf{.548}\\8.7/6154} & \shortstack{\textbf{.770}\\49.4/728} & \shortstack{\textbf{.849}\\69.7/655} & \shortstack{\textbf{.868}\\75.6/424}\\
\midrule
\multirow{5}{*}{\sav}
 & \raisebox{0.5\height}{Vanilla} & \shortstack{.465\\9.1/2317} & \shortstack{.468\\3.7/8543} & \shortstack{.506\\32.2/1327} & \shortstack{.613\\65.8/859} & \shortstack{.639\\74.1/501}\\
 \cmidrule{2-7}
 & \raisebox{0.5\height}{SparseViT~\cite{sparsevit}}  & \shortstack{.353\\4.6/3848} & \shortstack{.412\\6.3/8543} & \shortstack{.189\\18.4/3326} & \shortstack{.417\\64.5/1731} & \shortstack{.417\\34.8/883}\\
 \cmidrule{2-7}
 & \raisebox{0.5\height}{DynamicViT~\cite{tokenprune-imagenet}} & \shortstack{.210\\5.9/3849} & \shortstack{.307\\4.2/8543} & \shortstack{.181\\38.1/3327} & \shortstack{.286\\66.0/1731} & \shortstack{.417\\35.0/883}\\
 \cmidrule{2-7}
 & \raisebox{0.5\height}{VLTP~\cite{vltp}}       & \shortstack{.111\\9.5/4429} & \shortstack{.259\\3.3/9904} & \shortstack{.059\\37.3/3327} & \shortstack{.214\\72.5/1731} & \shortstack{.237\\33.9/883}\\
 \cmidrule{2-7}
 & \raisebox{0.5\height}{\textbf{\method}} & \shortstack{\textbf{.506}\\10.8/2243} & \shortstack{\textbf{.429}\\6.3/8450} & \shortstack{\textbf{.502}\\50.5/739} & \shortstack{\textbf{.541}\\71.9/645} & \shortstack{\textbf{.596}\\78.0/428}\\
\bottomrule
\end{tabular}
\caption{\textbf{Comparison against efficient token-reduction baselines.} \Jf{} (top) / \fps{}-\vram{} in MB (bottom) for each method-backbone pair. All baselines are image-domain token-reduction methods with no object- or video-level prior.}
\label{tab:baselines}
\end{table*}

\section{Ablation}
\label{sec:ablation}

We isolate the three knobs that define \method: \emph{(i)} which of the
two prunes does the work (read-side vs.\ write-side), \emph{(ii)} how
aggressive the read-side keep ratio $\rho$ is, and \emph{(iii)} how much
of the stored bank the write-side dilation $D_w$ retains. Throughout we explain each effect
averaged across several benchmarks.

\subsection{Read-side vs.\ write-side across architectures}
\label{sec:abl-readwrite}
Table~\ref{tab:abl-rw} decomposes \method into its two interventions
\emph{across every architecture in our suite}. For each one we report
Vanilla (V), \emph{read-side only} (R), and the full recipe (R$+$W, both prunes active).

\begin{table}[h]
\setlength{\abovecaptionskip}{3pt}
\centering\small
\setlength{\tabcolsep}{4pt}
\begin{tabular}{l|l|ccc}
\toprule
Model & Config & \Jf{} & \fps & \vram{} (MB) \\
\midrule
\multirow{3}{*}{STM~\cite{stm}}
  & V       & 0.802 & 37.6 & 1043 \\
  & R       & {0.735} & 40.4 & 873 \\
  & R$+$W   & {0.770} & \textbf{49.4} & \textbf{728} \\
\midrule
\multirow{3}{*}{XMem~\cite{xmem}}
  & V       & 0.862 & 37.2 & 845 \\
  & R       & 0.851 & 56.1 & 698 \\
  & R$+$W   & {0.849} & \textbf{69.7} & \textbf{655} \\
\midrule
\multirow{3}{*}{Cutie~\cite{cutie}}
  & V       & 0.879 & 23.8 & 436 \\
  & R       & 0.867 & 59.5 & 435 \\
  & R$+$W   & {0.868} & \textbf{75.6} & \textbf{424} \\
\midrule
\multirow{3}{*}{SAM2~\cite{sam2}}
  & V       & 0.882 & 9.6  & 1809 \\
  & R       & 0.812 & 10.2 & 1805 \\
  & R$+$W   & {0.797} & \textbf{11.4} & \textbf{1791} \\
\bottomrule
\end{tabular}
\caption{\textbf{Read-side vs.\ write-side ablation on \davis{}~val.} R = read-side
prune only; R+W = full \method{}. On XMem and Cutie, the read-side prune
delivers most of the speedup; adding the write-side prune yields a
further $\sim$25\% FPS gain and a further VRAM reduction over the
read-side-only row (6\% on XMem, 2.5\% on Cutie) at negligible \Jf{}
cost. On SAM2 and STM, the write-side prune acts as an
accuracy/efficiency knob: on SAM2 it trades 1.5 \Jf{} points for
+12\% FPS; on STM it improves \Jf{} by 3.5 points while delivering the
largest VRAM reduction in the suite (30.2\% relative to Vanilla).}
\label{tab:abl-rw}
\end{table}

The two prunes act on two different resources, and the decomposition
makes their division of labour explicit. The \emph{read-side} prune is
the \emph{speed} lever: by capping how many queries reach memory
attention each frame, it accounts for essentially all of \method's
throughput gain. The \emph{write-side} prune is the \emph{memory} lever:
by shrinking the stored bank it delivers most of the \vram{} reduction.
Stacking the two is close to free on any architectures with a top-$k$
readout filter --- the positions the write side discards
were already receiving near-zero attention weight at read time, so
dropping them from the bank perturbs the readout only negligibly.

\subsection{Keep ratio $\rho$ within the spatial prior}
\label{sec:abl-rho}
The read-side prune keeps the top-$\rho\!\cdot\!mn$ tokens \emph{inside}
the geometric prior, ranked by visual energy (Eq.~\ref{eq:read-keepset}),
so $\rho$ is the upper bound on how many queries reach memory attention
each frame. Table~\ref{tab:abl-rho} sweeps
$\rho \in \{0.2, 0.3, 0.5, 0.7\}$ on XMem DAVIS-17 with every
other component of \method held fixed.

\begin{table}[t]
\setlength{\abovecaptionskip}{3pt}
\centering\small
\setlength{\tabcolsep}{4pt}
\begin{tabular}{c|ccc|c}
\toprule
$\rho$ & \Jf{} & \fps & \vram{} (MB) & $|\mathcal{K}_r^{(t)}|_\mathrm{max}$ \\
\midrule
$0.2$  & 0.846 & \textbf{74.0} & \textbf{648} & 819  \\
$0.3$  & \textbf{0.849} & 69.7 & 655 & 1228 \\
$0.5$  & 0.852 & 58.4 & 681 & 2048 \\
$0.7$  & 0.853 & 49.1 & 712 & 2867 \\
\midrule
Vanilla ($\rho{=}1.0$) & 0.862 & 37.2 & 845 & 4096 \\
\bottomrule
\end{tabular}
\caption{\textbf{Keep-ratio sweep on XMem \cite{xmem} (\davis})
$\mathcal{J}\&\mathcal{F}$ stays within a $0.7$ point band
($0.846$--$0.853$) across $\rho\in[0.2,0.7]$, while \fps{}
drops by $1.5\times$ and \vram{} grows by $10\%$ as $\rho$
increases. The geometric prior already confines $\mathcal{K}_r$
to the cells relevant to attention readout; higher $\rho$ admits
only background tokens. We adopt $\rho{=}0.3$ throughout as the
point that maximises efficiency at near-optimal
$\mathcal{J}\&\mathcal{F}$.}
\label{tab:abl-rho}
\end{table}

The sweep is the empirical analogue of the Cauchy--Schwarz argument
\cite{steele2004cauchy} of Sec.~\ref{sec:why}: the geometric prior
already isolates the cells whose tokens carry the high $\ell_2$ energy,
and ranking by energy inside that small set converges on the
\Jf{}-relevant positions almost immediately. Admitting more tokens past
the prior only feeds the readout extra background energy, which its
softmax then normalises away --- leaving \Jf{} unchanged while still
paying the cross-attention cost. In other words, \method's efficiency
knob is essentially axis-aligned with $\rho$: the most aggressive budget
gives the best \fps{}/\vram{} at no measurable accuracy cost, which is why
we adopt $\rho{=}0.3$ as the deployment default.

\subsection{Write-side dilation on SAM2}
\label{sec:sam2-tradeoff}
Table~\ref{tab:abl-sam2} sweeps the write-side dilation $D_w$ on SAM2. The result is a clean accuracy--efficiency tradeoff
parameterised by $D_w$: a smaller $D_w$ aggressively shrinks the bank for
more FPS at the cost of more \Jf{}; a larger $D_w$ keeps more of the bank
intact and trades that FPS back. We adopt $D_w{=}24$ as the default configuration in our main results reported in Table~\ref{tab:main}.

\begin{table}[t]
\setlength{\abovecaptionskip}{3pt}
\centering\small
\begin{tabular}{l|c|ccc}
\toprule
Config & $D_w$ & $\mathcal{J}\&\mathcal{F}$ & \fps & \vram{} (MB) \\
\midrule
Vanilla SAM2              & ---  & 0.882 & 9.6  & 1809 \\
Read-side only            & ---  & 0.812 & 10.2 & 1805 \\
+ write-side (aggressive) & 12   & 0.785 & 13.6 & 1772 \\
+ write-side (default)    & 24   & 0.797 & 11.4 & 1791 \\
\bottomrule
\end{tabular}
\caption{\textbf{Write-side dilation sweep on SAM2 \cite{sam2} (\davis
\texttt{val}).} Larger $D_w$ keeps more positions per stored frame and
trades FPS for $\mathcal{J}\&\mathcal{F}$. The $D_w{=}24$ row is the configuration used in Table~\ref{tab:main}.}
\label{tab:abl-sam2}
\end{table}

\section{Discussion}
\label{sec:discussion}

\paragraph{One prune for speed, one for memory.}
The decomposition in Table~\ref{tab:abl-rw} gives \method a clean mental
model. The \emph{read-side} prune is the speed lever --- it caps how many
queries reach memory attention each frame and so carries essentially all
of the throughput gain --- while the \emph{write-side} prune is the
memory lever, shrinking the stored bank and delivering most of the
\vram{} reduction. Because the two act on different resources, their
benefits add: averaged over its five benchmarks, \method
gives $+58\%$\fps{}/$-32\%$\vram{} on STM,
$+26\%$\fps{}/$-16\%$\vram{} on XMem,
$+60\%$\fps{}/$-7\%$\vram{} on Cutie,
$+28\%$\fps{}/$-4\%$\vram{} on SAM2, and
$+22\%$\fps{}/$-6\%$\vram{} on SAM3.1 --- an overall
$+38.8\%$\fps{},$-13.1\%$\vram{} with no retraining and no change to the
released checkpoints.

\paragraph{Why pruning can \emph{improve} accuracy.} The reason is structural, not a free lunch from extra
capacity: by keeping only the geometrically-supported foreground tokens,
the read-side prune withholds the background context that a full-softmax
readout would otherwise integrate as noise. (as evident from the \better{Green} numbers in table \ref{tab:main})

\paragraph{Relation to query-only token pruning.}
Prior attention-time token pruning
\cite{vltp,tokenprune-imagenet,sparsevit} compresses only the queries
that read from the bank; it cannot shrink $N_\mathrm{mem}$. The \emph{write-side} prune is the lever that compresses $N_\mathrm{mem}$ and therefore the peak
memory of the long-video regime. The two prunes are complementary.

\section{Limitation}
\method holds the write keep-set fixed at the seed-frame footprint
dilated by $D_w$. On videos whose objects travel far outside that
footprint, the dilation either covers the trajectory at the cost of
\fps{} or undercovers it at the cost of \Jf{} --- the honest sensitivity
behind the long-video drops on the top-$k$ backbones. A learned,
motion-history-aware write keep-set would close this corner without
disturbing the rest of the recipe. However, in our case, we have a fixed keep-set in memory for all object occurrences throughout the video; hence, one object completely moving out of frame won't disrupt outputting masks for other objects. Likewise, the read-side keep-set is
propagated from the previous prediction; under catastrophic prediction
loss the streak-counted bounding-box fallback (Eq.~\ref{eq:recovery})
carries the system, but a dedicated re-detection branch would shorten the
recovery window further. Both are plug-in extensions, not re-designs.

\section{Conclusion}
\label{sec:conclusion}

We argued that the deployment ceiling of modern memory-bank VOS is not
the algorithmic gap between the strongest networks, but the unbounded
growth of the token budget that flows through their memory pipelines.
\method compresses that budget at the two structurally distinct points
where it accumulates --- between the image encoder and memory attention
(the read side, our speed lever) and between the memory encoder and the
memory bank (the write side, our memory lever) --- with a single,
training-free, geometric recipe that runs atop any VOS network and leaves the network's consolidation machinery intact.
\method retains far more accuracy than other efficient-VOS methods at lower time and memory budget across the benchmark datasets. The recipe ports to any
memory-bank VOS network that exposes the two encoder boundaries we
identify, turning the token budget of these systems from an architectural
constant into a tunable knob.

\paragraph{Acknowledgement}
The first author was affiliated with Attentive AI as a research engineer during the course of this work. We thank IIT Delhi for providing compute resources.

\bibliography{egbib}

\section*{Appendix}

\subsection*{Algorithm at a glance}
\label{sec:algo}

Algorithm~\ref{alg:rs3prune} collects the full pipeline as it runs
on a single video. No step introduces a new learned parameter and
no step requires backbone retraining: every
line either reuses a tensor already produced by the frozen VOS
network ($Z_t$, $\mathbf{q}_t$, $\hat{M}_t|_o$) or applies a
deterministic geometric operation (dilation, downsampling, top-$k$
selection) to it.

Per-frame cost follows directly from the keep-set sizes rather than
the full token grid. The read side attends
$|\mathcal{K}_r^{(t)}| \le \rho\, h w$ queries against the memory
bank instead of the full $hw$, so read-side attention cost scales
by $\rho$ relative to the dense pipeline. The write side commits
$|\mathcal{K}_w| \ll hw$ tokens per frame rather than the full
grid, so the memory bank after $T$ frames holds
$\mathcal{O}(T \cdot |\mathcal{K}_w|)$ entries instead of
$\mathcal{O}(T \cdot hw)$ --- the asymptotic source of the \vram{}
reduction reported in Experiments section of main paper. Both keep-sets are
recomputed by cheap geometric operations (bounding-box dilation,
nearest-neighbour downsampling, a single top-$k$) whose combined
cost is negligible next to a single memory-attention layer.

Two invariants hold throughout Algorithm~\ref{alg:rs3prune} and are
what make the recipe safe to drop into an unmodified backbone. First,
$\mathcal{K}_w$ is fixed once at $\tau_o$ and reused for every write
of the same sequence (line 6 is never re-entered), so the memory
bank's own consolidation and pruning logic --- which assumes a
constant per-frame stride --- continues to operate exactly as
designed. Second, whenever $D_w \ge r_s + c$, the containment
$\mathcal{P}^{(t)} \subseteq \mathcal{K}_w$ holds at every $t$, guaranteeing that the read side never
queries a bank position that was never written --- the read and
write prunes never work at cross purposes.

\begin{algorithm}[H]
\small
\DontPrintSemicolon
\SetAlgoLined
\SetKwInOut{Input}{Input}
\SetKwInOut{Output}{Output}
\SetKwInOut{Hyper}{Hyper-params}

\Input{Frames $I_0,\dots,I_{T-1}$; seed masks
       $\{M_o^{(0)}\}_{o\in\mathcal{O}}$ (palette PNG)}
\Hyper{$\rho{=}0.3$, $s{=}16$, $h{=}w{=}64$, $r_s{=}4$, $c{=}14$,
       $D_w$, closure kernel $9{\times}9$.}
\Output{Per-frame palette masks $\hat{M}_1,\dots,\hat{M}_{T-1}$.}
\BlankLine
\tcp{Frame $0$: seed prior, write keep-set, memory bank}
\If{text prompt $p$ given}{
  $M_o^{(0)} \leftarrow$ rasterised bounding box from a grounded
  detector on $(I_0, p)$\;
}
$\mathcal{B}_o^{(0)} \leftarrow \mathrm{bbox}(M_o^{(0)});\quad
 \hat{\mathcal{B}}_o \leftarrow \mathcal{B}_o^{(0)};\quad
 \Delta_o \leftarrow 0,\ \forall o$\;
$\mathcal{P}^{(0)} \leftarrow
  \bigcup_o \mathrm{dilate}_{r_s}(\mathrm{down}_s(\mathcal{B}_o^{(0)}))$
  \tcp*{Eq.~\ref{eq:seed-prior}}
$\mathcal{K}_w \leftarrow
  \bigcup_o \mathrm{dilate}_{D_w}(\mathrm{down}_s(M_o^{(\tau_o)}))$
  \tcp*{Eq.~\ref{eq:write-keepset}}
Seed SAM~2's memory bank from $\{M_o^{(0)}\}$
  via its native prompt path\;
\BlankLine
\tcp{Per-frame propagation}
\For{$t \leftarrow 1$ \KwTo $T-1$}{
  $Z_t \leftarrow \Phi_\mathrm{enc}(I_t)$
    \tcp*{$h{\times}w{\times}C$ tokens}
  \ForEach{$o \in \mathcal{O}$}{
    \eIf{$\hat{M}_{t-1}|_o$ non-empty}{
      $\mathcal{P}_o^{(t)} \leftarrow
        \mathrm{dilate}_{r_s}(\mathrm{down}_s(\hat{M}_{t-1}|_o))$
        \tcp*{Eq.~\ref{eq:dyn-prior}}
    }{
      $\mathcal{P}_o^{(t)} \leftarrow
        \mathrm{dilate}_{r_s+\min(\Delta_o,c)}(
          \mathrm{down}_s(\hat{\mathcal{B}}_o))$
        \tcp*{Eq.~\ref{eq:recovery}}
    }
  }
  $\mathcal{P}^{(t)} \leftarrow \bigcup_o \mathcal{P}_o^{(t)}$\;
  $\mathcal{K}_r^{(t)} \leftarrow
    \underset{i \in \mathcal{P}^{(t)}}{\mathrm{arg\text{-}top}_{\rho HW}}
    \, \|z_i\|_2^2$
    \tcp*{Eq.~\ref{eq:read-keepset}}
  $\tilde{\mathbf{q}}_t \leftarrow \mathbf{q}_t[\mathcal{K}_r^{(t)}];\
   \tilde{f} \leftarrow f[\mathcal{K}_r^{(t)}]$
    \tcp*{Eq.~\ref{eq:rope-gather}}
  $A \leftarrow
    \mathcal{S}_{\mathcal{K}_r^{(t)}}(\mathrm{softmax}(
      \tilde{\mathbf{q}}_t \mathbf{K}^\top/\sqrt{C_k})\,\mathbf{V})$
    \tcp*{Eq.~\ref{eq:read-attn}}
  $\{\hat{M}_t|_o\} \leftarrow \mathcal{D}(A);\
   \hat{M}_t|_o \leftarrow \mathrm{close}_{9\times9}(\hat{M}_t|_o),\,\forall o$
   \tcp*{Sec.~\ref{sec:morph}}
  \ForEach{$o \in \mathcal{O}$}{
    \eIf{$\hat{M}_t|_o$ non-empty}{
      $\hat{\mathcal{B}}_o \leftarrow \mathrm{bbox}(\hat{M}_t|_o);\
       \Delta_o \leftarrow 0$\;
    }{
      $\Delta_o \leftarrow \Delta_o + 1$\;
    }
  }
  $\hat{M}_t \leftarrow \mathrm{argmax}_o\, \hat{M}_t|_o$\;
  $\tilde{Z}_t \leftarrow Z_t[\mathcal{K}_w];\
   (\mathbf{K},\mathbf{V}) \leftarrow (\mathbf{K},\mathbf{V}) \cup
    \Phi_\mathrm{mem}(\tilde{Z}_t, \hat{M}_t)$
    \tcp*{sparse write}
  \textbf{emit} $\hat{M}_t$\;
}
\caption{\method --- one pass over a video.}
\label{alg:rs3prune}
\end{algorithm}

\end{document}